\documentclass{article}

     \usepackage[nonatbib,preprint]{neurips_2019}
\usepackage[utf8]{inputenc} 
\usepackage[T1]{fontenc}    
\usepackage{url}            
\usepackage{booktabs}       
\usepackage{amsfonts}       
\usepackage{nicefrac}       
\usepackage{enumitem}
\usepackage{epsfig}
\usepackage{amsmath}
\usepackage{amssymb}
\usepackage{mathtools}
\usepackage{times}
\usepackage{microtype}
\usepackage{graphicx}
\usepackage{caption}
\usepackage{subcaption}
\usepackage{float}
\usepackage{fancybox}
\usepackage{booktabs} 
\usepackage{xcolor}
\usepackage{hyperref}
\usepackage{cleveref}
\usepackage{algorithm}
\usepackage{algpseudocode}
\usepackage{xspace} 
\usepackage{scalerel,stackengine}
\usepackage{lineno}
\stackMath
\newcommand\reallywidehat[1]{%
\savestack{\tmpbox}{\stretchto{%
  \scaleto{%
    \scalerel*[\widthof{\ensuremath{#1}}]{\kern-.6pt\bigwedge\kern-.6pt}%
    {\rule[-\textheight/2]{1ex}{\textheight}}
  }{\textheight}%
}{0.5ex}}%
\stackon[1pt]{#1}{\tmpbox}%
}
\newcommand\Myperm[2][^n]{\prescript{#1\mkern-2.5mu}{}P_{#2}}

\newcommand{\tb}{\textbf}

\newcommand{\basealgo}{DEMINE\@\xspace}
\newcommand{\metaalgo}{Meta-DEMINE\@\xspace}
\newcommand{\secref}[1]{\S\ref{#1}}

\usepackage{amsfonts}
\usepackage{centernot}
\usepackage{xstring}
\usepackage{fixmath}
\usepackage{amsmath,amssymb,mathtools,bm,etoolbox}
\usepackage{ dsfont }
\usepackage{ upgreek }

\usepackage{grffile}
\newcommand{\VG}[1]{\mathbold {#1}}

\newcommand{\dimension}[2]{${#1} \in \mathbb{R}^{#2}$}

\DeclarePairedDelimiterX{\infdivx}[2]{(}{)}{%
  #1\;\delimsize\|\;#2%
}

\newcommand{\qed}{\hfill\blacksquare}

\DeclareMathOperator*{\argmax}{arg\,max}
\DeclareMathOperator*{\argmin}{arg\,min}

\newcommand{\FMRI}					{\ensuremath{\VG{\mathcal{X}}}}
\newcommand{\T}						{\ensuremath{T}}

\newcommand{\xV}					{\ensuremath{V_i}}
\newcommand{\FMRIdimFlat}					{\ensuremath{\xV \times \T}}

\begin{document}

\title{Data-Efficient Mutual Information Neural Estimator}

\author{
Xiao Lin$^1$\thanks{equal contribution} , Indranil Sur$^{1*}$, Samuel A. Nastase$^{2}$,\\
\textbf{Ajay Divakaran$^{1}$,  Uri Hasson$^{2}$ and Mohamed R. Amer$^{1}$}\\
\normalsize{$^{1}$SRI International, Princeton, NJ, USA}\quad \normalsize{$^{2}$Princeton University, Princeton, NJ, USA}
}
\date{}
\maketitle
%
%
%
\begin{abstract}
Measuring Mutual Information (MI) between high-dimensional, continuous, random variables from observed samples has wide theoretical and practical applications. Recent work, \textit{MINE}~\cite{MINE}, focused on estimating tight variational lower bounds of MI using neural networks, but assumed unlimited supply of samples to prevent overfitting. In real world applications, data is not always available at a surplus. In this work, we focus on improving data efficiency and propose a Data-Efficient MINE Estimator (DEMINE), by developing a relaxed predictive MI lower bound that can be estimated at higher data efficiency by orders of magnitudes. The predictive MI lower bound also enables us to develop a new meta-learning approach using task augmentation, \metaalgo, to improve generalization of the network and further boost estimation accuracy empirically. With improved data-efficiency, our estimators enables statistical testing of dependency at practical dataset sizes. We demonstrate the effectiveness of our estimators on synthetic benchmarks and a real world fMRI data, with application of inter-subject correlation analysis.
\end{abstract}
\section{Introduction}
Mutual Information (MI) is an important, theoretically grounded, measure of similarity between random variables. MI captures general, non-linear, statistical dependencies between random variables. It is a widely used quantity in various machine learning tasks ranging from classification to feature selection and neural network analysis.

A widely used approach for estimating MI from samples is using k-NN estimates, notably the KSG estimator~\cite{Kraskov2004}. \cite{Gao2017} provided a comprehensive review and studied the consistency and of asymptotic confidence bound of the KSG estimator \cite{Gao2018}. MI estimation can also be achieved by estimating individual entropy terms involved through kernel density estimation~\cite{Ahmad1976} or cross-entropy \cite{McAllester2018}. Overfitting can be reduced through partitioning the samples into different folds for modeling and for estimation. Despite of their fast and accurate estimations on random variables with few dimensions, MI estimation on high-dimensional random variables remains challenging for commonly used Gaussian kernels. Fundamentally, estimating MI requires the ability to accurately model the random variables, where high-capacity neural networks have shown excellent performance on complex high-dimensional signals such as text, image and audio. 

Recent works on MI estimation have focused on developing tight variational MI lower bounds where neural networks are used for signal modeling. The IM algorithm \cite{Agakov2004} introduces a variational MI lower bound, where a neural network $q(z|x)$ is learned as a variational approximation to the conditional distribution $P(Z|X)$. The IM algorithm requires the entropy, $H(Z)$, and $E_{XZ} \log q(z|x)$ to be tractable, which applies to latent codes of Variational Autoencoders (VAEs) and Generative Adversarial Networks (GANs) as well as categorical variables. \cite{MINE} introduces MI lower bounds \textit{MINE} and \textit{MINE-f} which allow the modeling of general random variables and shows improved accuracy for high-dimensional random variables, with application to improving generative models. \cite{Poole_2018} introduces a spectrum of energy-based MI estimators based on \textit{MINE} and \textit{MINE-f} lower bounds and a new TCPC estimator for the case when multiple samples from $P(Z|X)$ can be drawn. 

An important challenge that previous works overlooked is MI estimation using limited data. As the high-capacity neural networks tend to overfit. Variational estimators, such as \textit{MINE}, expect an impractically large number of samples to overcome overfitting and to reach high confidence. In addition, tighter lower bounds may also require more data to estimate. When limited number of samples are provided, estimations can suffer from high variance observed in \cite{Poole_2018}.  

To address the data efficiency challenge, our estimator, \basealgo, introduces predictive mode and meta-learning to the \textit{MINE} estimator family to greatly improve sample efficiency. We develop a relaxed, predictive variational lower bound based on \textit{MINE} that prevents overfitting by explicitly partitioning samples into training and validation. Furthermore, a predictive formulation allows us to incorporate techniques that improves generalization beyond curve fitting such as meta-learning. With these improvements, we show that \basealgo enables practical statistical testing of dependency in not only synthetic datasets but also for real world functional Magnetic Resonance Imaging (fMRI) data analysis for capturing nonlinear and higher-order brain-to-brain coupling.

An additional component to enhance our estimators is meta-learning. Meta-learning, or "learning to learn", seeks to improve the generalization capability of neural networks by searching for better hyper parameters~\cite{Maclaurin2015}, network architectures~\cite{Pham2018}, initialization~\cite{MAML,Finn2018,Kim2018} and distance metrics~\cite{Vinyals2016,Snell2017}. Meta-learning approaches have shown significant performance improvements in applications such as automatic neural architecture search~\cite{Pham2018}, few-shot image recognition~\cite{MAML} and imitation learning~\cite{Finn2017_imitation}. 

In particular, our estimator benefits from the Model-Agnostic Meta-Learning (MAML)~\cite{MAML} framework which is designed to improve few-shot learning performance. A network initialization is learned to maximize its performance when fine-tuned on few-shot learning tasks. Applications include few-shot image classification and navigation. We leverage the model-agnostic nature of MAML for MI estimation between generic random variable and adopt MAML for maximizing MI lower bounds. To construct a collection of diverse tasks for MAML learning from limited samples, inspired by MI's invariance to invertible transformations, we propose a task-augmentation protocol to automatically construct tasks by sampling random transformations to transform the samples. Results show reduced overfitting and improved generalization.


Our contributions are summarized as follows: 1) Data Efficient Mutual Information Neural Estimator (\basealgo); 2) New formulation of meta-learning using Task Augmentation (\metaalgo); 3) Application to real life, data scarse application (fMRI).
\section{Background}\label{sec:Background}
In this section, we will provide the background necessary to understand our approach\footnote{We follow the same notation in \cite{MINE}. We encourage the review of \cite{MINE, Poole_2018} to understand $I_{\text{MINE}}, I_{\text{EB1}}, \text{and } I_{\text{EB}}$.}. We define $X$ and $Z$ to be two random variables, $P(X,Z)$ is the joint distribution, and $P(X)$ and $P(Z)$ are the marginal distributions over $X$ and $Z$ respectively. Our goal is to estimate MI, $I(X;Z)$ given \textit{i.i.d.} sample pairs $(x_i,z_i)$, $i=1,2\dots n$ from $P(X,Z)$. Let $\mathcal{F}=\{T_\theta(x,z)\}_{\theta \in \Theta}$ be a class of scalar functions, where $\theta$ is the set of model parameters. Let $q(x|z)=p(x)\frac{e^{T_\theta(x,z)}}{\mathbb{E}_{(x,z)\sim P_{XZ}} e^{T_\theta(x,z)}}$. the following energy-based family of lower bounds of MI hold for any $\theta$:
\begin{equation}
\small
\begin{array}{rcl}
I(X;Z) & \ge &\mathbb{E}_{(x,z)\sim P_{XZ}} \log \frac{q(x|z)}{p(x)} = \mathbb{E}_{(x,z)\sim P_{XZ}} T_\theta(x,z) -  \mathbb{E}_{x\sim P_X}\log\mathbb{E}_{z\sim P_Z}  e^{T_\theta(x,z)} \triangleq I_{\text{EB1}} ~\cite{Poole_2018} \\ 
& \ge& \mathbb{E}_{(x,z)\sim P_{XZ}} T_\theta(x,z) - \log \mathbb{E}_{x\sim P_X, z\sim P_Z} e^{T_\theta(x,z)} \triangleq I_{\text{MINE}} ~\cite{MINE} \\ 
& \ge& \mathbb{E}_{(x,z)\sim P_{XZ}} T_\theta(x,z) -\mathbb{E}_{x\sim P_X,z\sim P_Z}  e^{T_\theta(x,z)} + 1 \triangleq I_{\text{MINE-f}}~\cite{MINE}, I_{\text{EB}}~\cite{Poole_2018} 
\end{array}
\label{eqn:lower_bounds}
\end{equation}
%
%
where, $\mathbb{E}$ is the expectation over the given distribution. 
%
%
%
%
%
%
Based on $I_{\text{MINE}}$, the \textit{MINE} estimator $\reallywidehat{I(X,Z)}_n$ is defined as in Eq.\ref{eqn:MINEestimator}. Estimators for $I_{\text{EB1}}$, $I_{\text{MINE-f}}$ and $I_{\text{EB}}$ can be defined similarly. 
\begin{equation}
\small
\reallywidehat{I(X;Z)}_n =  \sup_{\theta \in \Theta} \frac{1}{n} \sum_{i=1}^n T_\theta(x_i,z_i) - \log \frac{1}{n^2} \sum_{i=1}^n\sum_{j=1}^n e^{T_\theta(x_i,z_j)}.
\label{eqn:MINEestimator}
\end{equation}
With infinite samples to approximate expectation, Eq.\ref{eqn:MINEestimator} converges to the lower bound $\reallywidehat{I(X,Z)}_\infty=\sup_{\theta \in \Theta} I_{\text{MINE}}$.
%
%
%
Note that the number of samples $n$ needs to be substantially more than the number of model parameters $d=|\theta|$ to prevent $T_\theta(X,Y)$ from overfitting to the samples $(x_i,z_i)$, $i=1,2\dots n$ and overestimating MI. Formally, the sample complexity of \textit{MINE} is defined as the minimum number of samples $n$ in order to achieve Eq.\ref{eqn:samplecomplexity},
\begin{equation}
\small
    \text{Pr}(|\reallywidehat{I(X,Z)}_n-\reallywidehat{I(X,Z)}_\infty| \le \epsilon) \ge 1-\delta.
\label{eqn:samplecomplexity}
\end{equation}
Specifically, \textit{MINE} proves that under the following assumptions: 1) $T_\theta(X,Z)$ is $L$-Lipschitz; 2) $T_\theta(X,Z) \in [-M,M]$, 3) \{$\theta_i \in [-K,K], \quad\forall i\in{1,\ldots,d}$\}, the sample complexity of \textit{MINE} is given by Eq.\ref{eqn:MINEsamples}.
\begin{equation}
\small
    n \ge \frac{2M^2(d \log (16KL\sqrt{d}/\epsilon)+2dM+\log(2/\delta))}{\epsilon^2}.
\label{eqn:MINEsamples}
\end{equation}
%
%
%
For example, a neural network with dimension $d=10,000$, $M=1$, $K=0.1$ and $L=1$, achieving a confidence interval of $\epsilon=0.1$ with $95\%$ confidence would require $n \ge 18,756,256$ samples. This is achievable for synthetic example generated by Generative Adversarial Networks (GANs). For real data, however, the cost of data acquisition for reaching statistically significant estimation can be prohibitively expensive. 
We propose to use the MI lower bounds specified in Eq.\ref{eqn:lower_bounds} from a prediction perspective, inspired by cross-validation. Our estimator, \basealgo, improves sample complexity by disentangling data for lower bound estimation from data for learning a generalizable $T_\theta(X,Z)$. \basealgo enables high-confidence MI estimation on small datasets. Bound tightness is further improved by \metaalgo by using meta-learning to learn generalizable $T_\theta(X,Z)$. 
\section{Approach}
\secref{sec:Estimator} specifies \basealgo for predictive MI estimation and derives the confidence interval; \secref{sec:Meta} formulates \metaalgo, explains task augmentation, and defines the optimization algorithms.
\subsection{Predictive Mutual Information Estimation}\label{sec:Estimator}
In \basealgo, we interpret the estimation of \textit{MINE-$f$} lower bound\footnote{\textit{MINE} lower bound can also be interpreted in the predictive way, but will result in a higher sample complexity than \textit{MINE-$f$} lower bound. We choose \textit{MINE-f} in favor of a lower sample complexity over bound tightness.} Eq.\ref{eqn:lower_bounds} as a learning problem. The goal is to infer the optimal network $T_{\theta^*}(X,Z)$ with parameters $\theta^*$ using a limited number of samples defined as follows: 
\begin{equation*}
\small
    \theta^*=\argmax_{\theta \in \Theta} \mathbb{E}_{P_{XZ}} T_\theta(X,Z) - \mathbb{E}_{P_X} \mathbb{E}_{P_Z}  e^{T_\theta(X,Z)} +1.
\end{equation*}
Specifically, samples from $P(X,Z)$ are subdivided into a training set $\{(x_i,z_i)_{\text{train}}, i=1,\ldots,m\}$ and a validation set $\{(x_i,z_i)_{\text{val}}, i=1,\ldots,n$\}. The training set is used for learning a network $\tilde{\theta}$ as an approximation to $\theta^*$ whereas the validation set is used for computing the \basealgo estimation $\reallywidehat{I(X,Z)}_{n,\tilde{\theta}}$ defined as in Eq.\ref{eqn:OurEstimator}. 
\begin{equation}
\small
    \reallywidehat{I(X,Z)}_{n,\tilde{\theta}} = \frac{1}{n} \sum_{i=1}^n T_{\tilde{\theta}}(x_i,z_i)_{\text{val}} - \frac{1}{n^2} \sum_{i=1}^n\sum_{j=1}^n e^{T_{\tilde{\theta}}(x_i,z_j)_\text{val}} +1 
\label{eqn:OurEstimator}
\end{equation}
We propose a approache to learn $\tilde{\theta}$, \basealgo. \basealgo learns $\tilde{\theta}$ by maximizing the MI lower bound on the training set as follows:
\begin{equation*}
\small
\tilde{\theta}=\argmin_{\theta \in \Theta} \mathcal{L}(\{(x,z)\}_{train},\theta), \text{where,}
\end{equation*}
\begin{equation}
\small
    \mathcal{L}(\{(x,z)\}_\mathcal{B},\theta)= -\frac{1}{|\mathcal{B}|} \sum_{i=1}^{|\mathcal{B}|} T_{\theta}(x_i,z_i)_\mathcal{B} + \frac{1}{|\mathcal{B}|^2} \sum_{i=1}^{|\mathcal{B}|}\sum_{j=1}^{|\mathcal{B}|} e^{T_{\theta}(x_i,z_j)_\mathcal{B}}-1.
\label{eqn:loss}
\end{equation}
\noindent The \basealgo algorithm is shown in Algorithm~\ref{algo:basealgo}.
\begin{algorithm}[H]
\caption{\basealgo}\label{algo:basealgo}
\begin{algorithmic}[1]
\Statex \textbf{Input Data}: $\{(x,z)_\text{train}, (x,z)_{\text{val}}\}$
\Statex \textbf{Parameters}: Batch $\mathcal{B}$, Iterations $N_O$, Learning rate $\eta$ 
\Statex \textbf{Output}: MI, $T_\theta (X,Z)$
\State $\theta^{(0)} \leftarrow$ Xavier Initialization \cite{xavier}
\For{i = 1 : $N_O$}
    \State Sample a batch of ${(x_{i},z_{i})}_\mathcal{B} \sim (x,z)_{\text{train}}$
    \State Compute $\mathcal{L}\left( (x_{i},z_{i})_\mathcal{B},\theta^{(i-1)}\right)$
    \State Compute $\nabla_\theta^{(i)} \mathcal{L}$ -- gradient for $\theta$
    \State Update $\theta^{(i)}$ using Adam with $\eta$
\EndFor 
\State $\text{MI}=\reallywidehat{I(X,Z)}_{n,\theta^{(N_O)}}$ 
\State \textbf{return} MI, $\theta^{(N_O)}$
\end{algorithmic}
\end{algorithm}

\textbf{Sample complexity analysis.} 
%
Because $\tilde{\theta}$ is learned independently of validation samples $\{(x_i,z_i)_{\text{val}}, i=1,\ldots,n\}$, the sample complexity of the \basealgo estimator does not involve the model class $\mathcal{F}$ and the sample complexity is greatly reduced compared to \textit{MINE-f}. 
\basealgo estimates $\reallywidehat{I(X,Z)}_{\infty,\tilde{\theta}}$ when infinite number of samples are provided, defined as: 
%
%
%
\begin{equation}
\small
\begin{array}{rcl}
\reallywidehat{I(X,Z)}_{\infty,\tilde{\theta}} &=& \mathbb{E}_{P_{XZ}} T_{\tilde{\theta}}(X,Z) - \mathbb{E}_{P_X} \mathbb{E}_{P_Z}  e^{T_{\tilde{\theta}}(X,Z)} +1 \\
& \leq& \sup_{\theta \in \Theta} \mathbb{E}_{P_{XZ}} T_\theta(X,Z) - \mathbb{E}_{P_X} \mathbb{E}_{P_Z} e^{T_\theta(X,Z)} +1 \leq I(X;Z)
\end{array}
\label{eqn:tightBound}
\end{equation}
We now derive the sample complexity of \basealgo defined as the number of samples $n$ required for $\reallywidehat{I(X,Z)}_{n,\tilde{\theta}}$ to be a good approximation to $\reallywidehat{I(X,Z)}_{\infty,\tilde{\theta}}$ in Theorem 1.

\noindent\textbf{Theorem 1.} For $T_{\tilde{\theta}}(X,Z)$ bounded by $[L,U]$, given any accuracy $\epsilon$ and confidence $\delta$, we have:
\begin{equation*}
\small
\text{Pr}(|\reallywidehat{I(X,Z)}_{n,\tilde{\theta}}-\reallywidehat{I(X,Z)}_{\infty,\tilde{\theta}}| \le \epsilon) \ge 1-\delta 
\end{equation*}
when the number of validation samples $n$ satisfies:
\begin{equation}
\small
n \ge n^*, \text{\textit{s.t.}\,} f(n^*)\equiv\min_{0\le \xi \le \epsilon} 2e^{-\frac{2 \xi^2 n^*}{(U-L)^2}}+4e^{-\frac{(\epsilon-\xi)^2 n^*}{2(e^U-e^L)^2}} = \delta
\label{eqn:nsamples}
\end{equation}
\textbf{Proof.} Since $T_{\tilde{\theta}}(X,Z)$ is bounded by$[L,U]$, applying the Hoeffding inequality to the first half of Eq.\ref{eqn:OurEstimator}:
\begin{equation*}
\small
\text{Pr}(|\frac{1}{n} \sum_{i=1}^n T_{\tilde{\theta}}(x_i,z_i)-\mathbb{E}_{P_{XZ}} T_{\tilde{\theta}}(X,Z)| \ge \xi) \le 2e^{-\frac{2 \xi^2 n}{(U-L)^2}}
\end{equation*}
As $e^{T_\theta(X,Z)}$ is bounded by $[e^L,e^U]$, applying the Hoeffding inequality to the second half of Eq.\ref{eqn:OurEstimator}:
\begin{equation*}
\small
\begin{array}{lcl}
\text{Pr}(|\mathbb{E}_{P_X} \mathbb{E}_{P_Z}  e^{T_\theta(X,Z)}- \frac{1}{n} \sum_{i=1}^n \mathbb{E}_{P_Z}  e^{T_{\tilde{\theta}}(x_i,z)}| \ge \zeta) &\le& 2e^{-\frac{2 \zeta^2 n}{(e^U-e^L)^2}} \\
\text{Pr}(|\mathbb{E}_{P_Z} \frac{1}{n} \sum_{i=1}^n  e^{T_\theta(x_i,z)}-\frac{1}{n} \sum_{j=1}^n \frac{1}{n} \sum_{i=1}^n  e^{T_{\tilde{\theta}}(x_i,z_j)}| \ge \zeta) &\le& 2e^{-\frac{2 \zeta^2 n}{(e^U-e^L)^2}}
\end{array}
\end{equation*}
Combining the above bounds results in:
\begin{equation*}
\small
\text{Pr}(|\reallywidehat{I(X,Z)}_{n,\tilde{\theta}}-\reallywidehat{I(X,Z)}_{\infty,\tilde{\theta}}| \le \xi+2\zeta) \ge 1-2e^{-\frac{2 \xi^2 n}{(U-L)^2}}-4e^{-\frac{2 \zeta^2 n}{(e^U-e^L)^2}} 
\end{equation*}
By solving $\xi$ to minimize $n$ according to Eq.\ref{eqn:nsamples} we have:
\begin{equation*}
\small
\text{Pr}(|\reallywidehat{I(X,Z)}_{n,\tilde{\theta}}-\reallywidehat{I(X,Z)}_{\infty,\tilde{\theta}}| \le \epsilon) \ge 1-\delta. \quad\quad \qed
\end{equation*}
Compared to \textit{MINE}, as per the example shown in \secref{sec:Background}, for $M=1$ (\textit{i.e.} $L=-1$ and $U=1$), $\delta=0.05$, $\epsilon=0.1$, our estimator requires $n=10,742$ compared to \textit{MINE} requiring $n = 18,756,256$ \textit{i.i.d} validation samples to estimate a lower bound, 
which makes MI-based dependency analysis feasible for domains where data collection is prohibitively expensive, \textit{e.g.} fMRI brain scans. In practice, sample complexity can be further optimized by tuning hyperparameters $U$ and $L$.

Note that the sample complexity of our approach, \basealgo, for estimating Eq.\ref{eqn:tightBound} does not depend on network size $d$. The improved sample complexity seemingly comes at a cost of bound tightness guarantees. In fact, to guarantee bound tightness of Eq.\ref{eqn:tightBound}, $O(d \log d)$ examples would still be theoretically required to learn $\tilde{\theta}$ with guaranteed close values to $\theta^*$, and the total data cost would be on par with \textit{MINE}. In practice,such a learnability bound is known to be overly loose, as over-parameterized neural networks have been shown to generalize well in classification and regression tasks~\cite{Zhang_2016}. Fundamentally, what determines bound tightness is the generalization error of $\tilde{\theta}$ -- to which the learnability bound is serving as a proxy. Empirically, not only that the bound tightness of \basealgo is as good as \textit{MINE} so the loss of guaranteed tightness did not affect empirical tightness, but the learning-based formulation of \basealgo also allows further bound tightness improvements by learning $\tilde{\theta}$ that generalizes beyond curve fitting using meta-learning. 

In the following section, we present a meta-learning formulation, \metaalgo, that learns $\tilde{\theta}$ for generalization given the same model class and training samples.
\subsection{Meta-Learning}\label{sec:Meta}
Given training data $\{(x_i,z_i)_{\text{train}},i=1,\ldots m\}$, \metaalgo algorithm first generates MI estimation tasks each consisting of a meta-training split $\text{A}$ and a meta-val split $\text{B}$ through a novel \textit{task augmentation} process. A parameter initialization $\theta_\text{init}$ is then learned to maximize MI estimation performance on the generated tasks using initialization $\theta_\text{init}$ as shown in Eq.\ref{eqn:metaapproach_init}.
\begin{equation}
\small
    \theta_\text{init}=\argmin_{\theta^{(0)} \in \Theta} \mathbb{E}_{(\text{A},\text{B})\in \mathcal{T}} \mathcal{L}((x,z)_{\text{B}},\theta^{(t)}), \text{with },\theta^{(t)}\equiv\text{MetaTrain}\big((x,z)_{\text{A}},\theta^{(0)}\big).
\label{eqn:metaapproach_init}
\end{equation}
Here $\theta^{(t)}=\text{MetaTrain}\big((x,z)_{\text{A}},\theta^{(0)}\big)$ is the meta-training process of starting from an initialization $\theta^{(0)}$ and applying SGD\footnote{In practice, Adam~\cite{adam} is used for faster optimization. Illustrating SGD for simplicity. } over $t$ steps to learn $\theta$ where in every meta training iteration we have:
\begin{equation*}
\small
    \theta^{(t)} \leftarrow \theta^{(t-1)}-\gamma \nabla \mathcal{L}((x,z)_{\text{A}},\theta^{(t-1)}).
\end{equation*}
Finally, $\tilde{\theta}$ is learned using the entire training set $\{(x_i,z_i)_{\text{train}}, i=1,\ldots, m\}$ with $\theta_\text{init}$ as initialization:
\begin{equation*}
\small
    \tilde{\theta}=\text{MetaTrain}\big((x,z)_{\text{train}},\theta_\text{init}\big).
\end{equation*}
\noindent\textbf{Task Augmentation:} \metaalgo adapts MAML~\cite{MAML} for MI lower bound maximization. MAML has been shown to improve generalization performance in $N$-class $K$-shot image classification. MI estimation, however, does not come with predefined classes and tasks. A naive approach to produce tasks would be through cross validation -- partitioning training data into meta-training and meta-validation splits. However, merely using cross-validation tasks is prone to overfitting -- a $\theta_\text{init}$, which memorizes all training samples would as a result have memorized all meta-validation splits. Instead, \metaalgo generates tasks by augmenting the cross validation tasks through \textit{task augmentation}. Training samples are first split into meta-training and meta-validation splits, and then transformed using the same random invertible transformation to increase task diversity. \metaalgo generates invertible transformation by sequentially composing the following functions:
\begin{center}
\small
$\begin{array}{lll}
\textit{Mirror}:   & m(x) = (2n-1)x,                                     & n \sim \text{Bernoulli}(\frac{1}{2}),\\
\textit{Permute}: & P(x)=\Myperm{d},                                & \text{Permute dimensions.} \\
\textit{Offset}:  & O(x) = x+\epsilon,                              & \epsilon \sim \mathcal{U}(-0.1,0.1),\\
\textit{Gamma}:   & G(x) = \text{sign}(x)\left|x\right|^{\gamma},   & \gamma \sim \mathcal{U}(0.5,2),
\end{array}$
\end{center}
Since the MI between two random variables is invariant to invertible transformations on each variable, $\text{MetaTrain}$ is expected to arrive at the same MI lower bound estimation regardless of the transformation applied. At the same time, memorization is greatly suppressed, as the same pair $(x,z)$ can have different $\log \frac{p(x,z)}{p(x)p(z)}$ under different transformations. More sophisticated invertible transformations (affine, piece-wise linear) can also be added. Task augmentation is an orthogonal approach to data augmentation. Using image classification as an example, data augmentation generates variations of the image, translated, or rotated images assuming that they are valid examples of the class. Task augmentation on the other hand, does not make such assumption. Task augmentation requires the initial parameters $\theta_\text{init}$ to be capable of recognizing the same class in a world where all images are translated and/or rotated, with the assumption that the optimal initialization should easily adapt to both the upright world and the translated and/or rotated world. 

\noindent\textbf{Optimization:} Solving $\theta_\text{init}$ using the meta-learning formulation Eq.\ref{eqn:metaapproach_init} poses a challenging optimization problem. 
The commonly used approach is back propagation through time (BPTT) which computes second order gradients and directly back propagate gradient from $\text{MetaTrain}((x,z)_{\text{A}},\theta^{(0)})$ to  $\theta_\text{init}$. BPTT is very effective for a small number of optimization steps, but is vulnerable to exploding gradients and is memory intensive. In addition to BPTT, we find that stochastic finite difference algorithms such as Evolution Strategies (ES)~\cite{es} and Parameter-Exploring Policy Gradients (PEPG)~\cite{pepg} can sometimes improve optimization. In practice, we use BPTT or PEPG to optimize Eq.\ref{eqn:metaapproach_init} depending on the problem. \metaalgo algorithm is specified in Algorithm~\ref{algo:metaalgo}. 

\begin{algorithm}[t!]
\caption{\metaalgo}\label{algo:metaalgo}
\begin{algorithmic}[1]
\Statex \textbf{Input Data}: $\{(x,z)_{\text{train}}, (x,z)_{\text{val}}\}$
\Statex \textbf{Parameters}: batch $\mathcal{B}$, Meta Learning Iterations $N_M$, Task Augmentation Iterations $N_T$, Optimization Iterations $N_O$, Ratio $r$, Learning rate $\eta$, Meta Learning Rate $\eta_{\text{meta}}$
\Statex \textbf{Output}: MI, $T_{\theta_\text{init}} (X,Z)$, $T_\theta (X,Z)$
\For{i = 1 : $N_M$}
    \For{j = 1 : $N_T$}
        \State $A = r \times \text{train}, B = \text{train}-A$
        \State Split $(x,z)_{\text{train}}$ into $(x,z)_{\text{A}}$ and $(x,z)_{\text{B}}$
        \State Transformation $R_x$ for $x,$ $R_x(\cdot)=\text{m}(\text{P}(\text{O}(\text{G}(\cdot))))$
        \State Transformation $R_z$ for $z,$ $R_z(\cdot)=\text{m}(\text{P}(\text{O}(\text{G}(\cdot))))$
        \State $\theta_{\text{meta}}^{(0)}\leftarrow\theta_\text{init}$
        \For{k = 1 : $N_O$}
            \State Sample a batch of $(x,z)_{\mathcal{B}} \sim (x,z)_{\text{A}}$
            \State Compute $\mathcal{L}\big( (R_x(x),R_z(z))_{\mathcal{B}},\theta_{\text{meta}}^{(k)}\big)$ 
            \State Compute $\nabla_{\theta_{\text{meta}}^{(k)}}\mathcal{L}$ -- gradient for $\theta_{\text{meta}}$
            \State Update $\theta_{\text{meta}}$ using Adam \cite{adam} with $\eta$
        \EndFor
        \State Compute $\mathcal{L}_{\text{meta}}\big((R_x(x),R_z(z))_{\text{B}},\theta^{(N_O)}_{\text{meta}}\big)$ 
        \State Compute $\nabla_{\theta_{0}}\mathcal{L}_{\text{meta}}$ -- gradient to $\theta_\text{init}$ using BPTT
    \EndFor
    \State Update $\theta_\text{init}$ using Adam \cite{adam} with $\eta_{meta}$
\EndFor
\State $\theta^{(0)}\leftarrow\theta_\text{init}$
\For{i = 1 : $N_O$}
    \State Sample a batch of $(x,z)_{\mathcal{B}} \sim (x,z)_{\text{train}}$
    \State Compute $\mathcal{L}\big((x,z)_{\mathcal{B}},\theta^{(i)}\big)$ 
    \State Compute gradient $\nabla_{\theta} \mathcal{L}$
    \State Update $\theta$ using Adam with $\eta$
\EndFor
\State Compute $\text{MI}=\mathcal{L}\big((x,z)_{\text{val}},\theta^{(N_O)}\big)$
\State \textbf{return} MI, $\theta_\text{init}$, $\theta^{(N_O)}$
\end{algorithmic}
\end{algorithm}

\section{Evaluation on Synthetic Datasets}\label{sec:expsetup}
\textbf{Dataset.} We evaluate our approaches \basealgo and \metaalgo against baselines and state-of-the-art approaches on 3 synthetic datasets: 1D Gaussian, 20D Gaussian and sine wave. For 1D and 20D Gaussian datasets, following~\cite{MINE}, we define two $k$-dimensional multivariate Gaussian random variables $X$ and $Z$ which have component-wise correlation $corr(X_i,Z_j)=\delta_{ij}\rho$, where $\rho \in (-1,1)$ and $\delta_{ij}$ is Kronecker's delta. Mutual information $I(X;Z)$ has a closed form solution $I(X;Z)=-k\ln(1-\rho^2)$. For sine wave dataset, we define two random variables $X$ and $Z$, where $X \sim \mathcal{U}(-1,1)$, $Z=\sin (aX+\frac{\pi}{2})+0.05\epsilon$, and $\epsilon \sim \mathcal{N}(0,1)$. Estimating mutual information accurately given few pairs of $(X,Z)$ requires the ability to extrapolate the sine wave given few examples. Ground truth MI for sine wave dataset is approximated by running the the KSG Estimator~\cite{Kraskov2004} on $1,000,000$ samples.

\textbf{Implementation.} We compare our estimators, \basealgo and \metaalgo, against the KSG estimator~\cite{Kraskov2004} MI-KSG and MINE-f. For both \basealgo and \metaalgo, we study variance reduction mode, referred to as \textit{-vr}, where hyperparameters are selected by optimizing 95\% confident estimation mean ($\mu-2\sigma_\mu$) and statistical significance mode, referred to as \textit{-sig}, where hyperparameters are selected by optimizing 95\% confident MI lower bound ($\mu-\epsilon$). Samples $(x,z)$ are split into 50\%-50\% as $(x,z)_{\text{train}}$ and $(x,z)_{\text{val}}$. 

We use a separable network architecture $T_\theta(x,z)=M\big(\text{tanh}( w \cos\left< f(x),g(z)\right> + b ) -t\big)$. $f$ and $g$ are MLP encoders that embed signals $x$ and $z$ into vector embeddings. Hyperparameters $t\in [-1,1]$ and $M$ control upper and lower bounds $T_\theta(x,z) \in [-M(1+t),M(1-t)]$. Parameters $w$ and $b$ are learnable parameters. MLP design and optimization hyperparameters are selected using Bayesian hyperparameter optimization~\cite{hyperopt} with 3-fold cross-validation on $(x,z)_{\text{train}}$ over 1,000 iterations.

Hyperparameter search on DEMINE-vr and DEMINE-sig was conducted using the hyperopt package \footnote{Hyperopt package: \url{https://github.com/hyperopt/hyperopt}.}. Seven hyper parameters were involved in hyperparameter search: 1) number of encoder layers $[1,5]$, 2) encoder hidden size $[8,256]$, 3) learning rate $\eta$ $[10^{-4},3\times 10^{-1}]$ in log scale, 4) number of optimization iterations $N_O$ $[5,200]$ (sine wave $[5,5000]$) in log scale, 5) batch size $\mathcal{B}$ $[256,1024]$, 6) $M$, $[10^{-3},5]$ in log scale, 7) $t$, $[-1,1]$. Mean $\mu$ and sample standard deviation $\sigma$ of MI estiamte computed over 3 fold cross validation on $(x,z)_{\text{train}}$. DEMINE-vr maximizes two sigma low $\mu-2\sigma_{\mu}$ where $\sigma_{\mu}=\frac{1}{\sqrt{3}}\sigma$. DEMINE-sig maximizes statistical significance $\mu-\epsilon$ where $\epsilon$ is two-sided 95\% confidence interval of MI. Meta-DEMINE-vr and Meta-DEMINE-sig subsequently reuse these hyperparameters as DEMINE-vr and DEMINE-sig. 

Meta-learning hyperparameters are chosen as outer loop $N_M=3,000$ iterations, task augmentation $N_T=1$ iterations, $r=0.8$, $\eta_{\text{meta}}=\frac{\eta}{3}$, with task augmentation mode $m(P(O(\cdot)))$. $N_O$ capped at 30 iterations for 1D and 20D Gaussian datasets due to memory limit. The sine wave datasets require large $N_O$, we used PEPG~\cite{pepg} rather than BPTT.

For MI-KSG, we use off-the-shelf implementation~\cite{Gao2017} with default number of nearest neighbors k=3. MI-KSG does not provide any confidence interval. For MINE-f, we use the same network architecture same as DEMINE-vr. we implement both the original formulation which optimizes $T_\theta$ on $(x,z)$ till convergence (10k iters), as well as our own implementation MINE-f-ES with early stopping, where optimization is stopped after the same number of iterations as DEMINE-vr to control overfitting. 

\begin{figure}
\centering
\begin{minipage}{\textwidth}
\centering
  \subcaptionbox{20D Gaussian dataset, $N=300$ samples}{\includegraphics[width=0.48\linewidth]{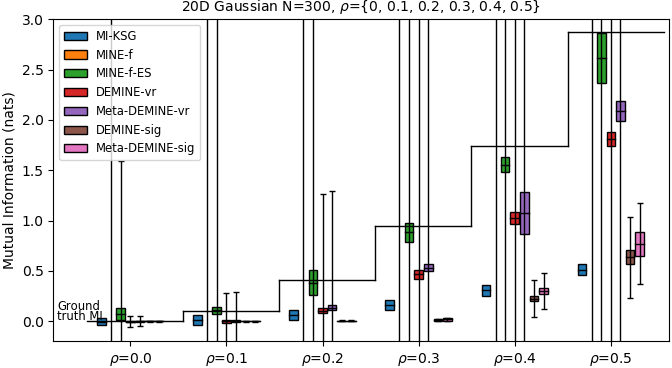}}
  \subcaptionbox{1D Gaussian dataset, $\rho=0.8$}{\includegraphics[width=0.48\linewidth]{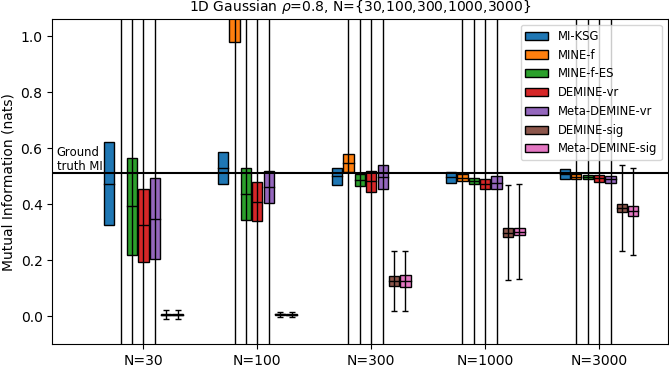}}\\
  \subcaptionbox{20D Gaussian dataset, $\rho=0.3$}{\includegraphics[width=0.48\linewidth]{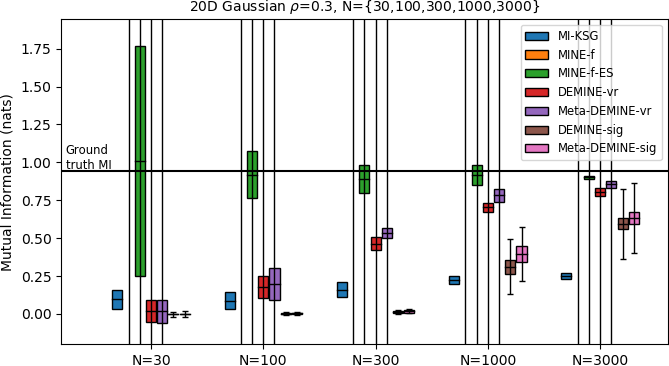}}
  \subcaptionbox{Sine wave dataset, $a=8\pi$}{\includegraphics[width=0.48\linewidth]{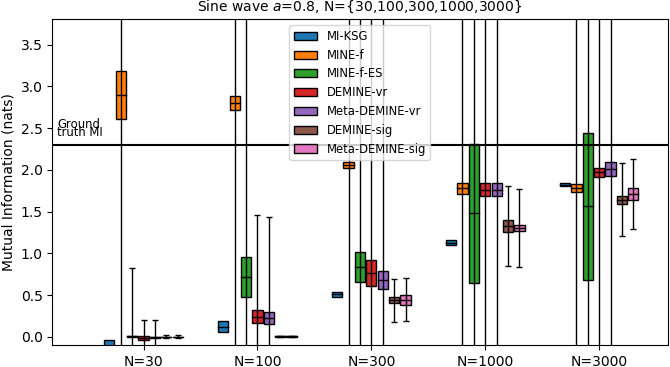}}
  \caption{Comparing MI Estimation performance of \basealgo and \metaalgo with the KSG estimator~\cite{Kraskov2004} and MINE-f~\cite{MINE} on different datasets using varying number of samples. The bars show estimator mean and standard deviation averaged over 5 runs with different seeds. The errorbars show 95\% confidence interval (not available for MI-KSG). The statistical significance focused variants \basealgo-sig and \metaalgo-sig achieves the highest 95\% confident MI estimation. \metaalgo improves over \basealgo most of the time. Best viewed in color.}
  \label{fig:main_results}
\end{minipage}
\vspace{10pt}
\begin{minipage}{\textwidth}
  \centering
  \subcaptionbox{\metaalgo-vr $N_O=0$.} {\includegraphics[width=0.32\linewidth]{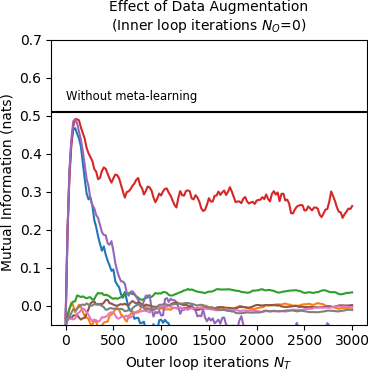}}
  \subcaptionbox{\metaalgo-vr $N_O=10$.} {\includegraphics[width=0.32\linewidth]{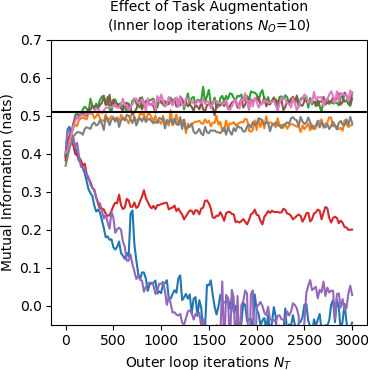}}
  \subcaptionbox{\metaalgo-vr $N_O=20$.} {\includegraphics[width=0.32\linewidth]{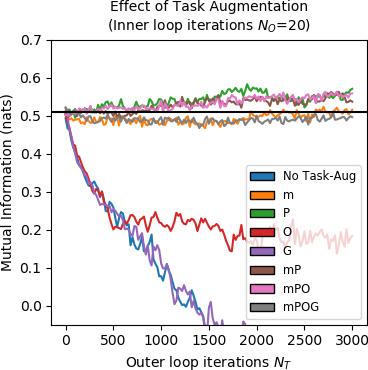}}
  \caption{To study the effect of task augmentation and number of adaptation steps, we run \metaalgo-vr with different task augmentation modes and vary number of adaptation iterations $N_O \in \{0,10,20\}$ on Gaussian 20D, $\rho=0.3$ dataset. Combinations of permutation and mirroring operations are effective in reducing overfitting and improving performance. Best viewed in color.}
  \label{fig:task_aug}
\end{minipage}
\vspace{10pt}
\begin{minipage}{.33\linewidth}
    \scriptsize
    \tabcolsep=0.11cm
    \captionof{table}{Number of HCP-MMP1 regions with significant pairwise correlation (r) and MI (\basealgo, \metaalgo) during listening.}
    \centering
    \label{table:fmri_sig}
    \begin{tabular}{ l |c c c  }
    \toprule
    No. shared & r & \basealgo & Meta  \\
     &  &  & -DEMINE  \\
    \midrule
    r & 37 & 24 & 23  \\
    DEMINE & 24 & 28 & 26  \\
    Meta-DEMINE & 23 & 26 & 29  \\
    \bottomrule
    \end{tabular}
\end{minipage}
\hfill
\begin{minipage}{.63\linewidth}
    \scriptsize
    \tabcolsep=0.11cm
    \captionof{table}{Segment classification accuracy for NeuralMI versus Pearson's correlation in 1-vs-1 and 1-vs-rest*.}
    \centering
    \label{table:fmri_classification}
    \begin{tabular}{ l |c  c  c  c | c | c  c  c  c | c  }
    \toprule
    Classification	& \multicolumn{5}{c}{\textbf{ISC Mask}} & \multicolumn{5}{c}{\textbf{dDMN Mask}}   \\
    Accuracy (\%) & P & F & Br & Bk & MI & P & F & Br & Bk & MI  \\
    \midrule
    Chance & 3.7 & 1.8 & 2.6 & 1.9 & N/A & 3.7 & 1.8 & 2.6 & 1.9 & N/A  \\
    Pearson's r 1vR & 35.0 & 20.4 & 25.8 & 31.5 & N/A & 14.8 &  6.4 & \tb{11.8} & 9.9 & N/A  \\
    \basealgo 1vR    & 42.8 & 28.0 & 32.8 & 35.9 & 0.637 & \tb{16.5} &  \tb{7.9} & 11.6 & \tb{12.0} & \tb{0.035}  \\
    \metaalgo 1vR  &  \tb{47.2}  & \tb{32.5} & \tb{39.9} & \tb{41.0} & \tb{0.752} & 13.7 & 7.9 & 8.2 & 8.9 & 0.031 \\
    \bottomrule
    \end{tabular}
    Abbreviations: P: Pieman; F: Forgotten; Br: Bronx; Bk: Black, MI: Mutual Information.\\
    *Note that all the results are averaging over other subjects.
\end{minipage}
\end{figure}

\paragraph{Results.} Figure~\ref{fig:main_results}(a) shows MI estimation performance on 20D Gaussian datasets with varying $\rho \in \{ 0,0.1,0.2,0.3,0.4,0.5\}$ using $N=300$ samples. Results are averaged over 5 runs to compare estimator bias, variance and confidence. Note that \metaalgo-sig detects the highest $p<0.05$ confidence MI, outperforming \basealgo-sig which is a close second. Both detect $p<0.05$ statistically significant dependency starting $\rho=0.3$, whereas estimations of all other approaches are low confidence. It shows that in contrary to common belief, estimating the variational lower bounds with high confidence can be challenging under limited data. MINE-f estimates MI$>3.0$ and MINE-f-ES estimates positive MI when $\rho=0$, both due to overfitting, despite MINE-f-ES having the lowest empirical bias. \basealgo variants have relatively high empirical bias but low variance due to tight upper and lower bound control, which provides a different angle to understand bias-variance trade off in MI estimation~\cite{Poole_2018}.

Figure~\ref{fig:main_results}(b,c,d) shows MI estimation performance on 1D, 20D Gaussian and sine wave datasets with fixed $\rho=0.8,0.3$ and $a=8\pi$ respectively, with varying $N \in \{30,100,300,1000,3000\}$ number of samples. More samples asymptotically improves empirical bias across all estimators. As opposed to 1D Gaussian datasets which are well solved by $N=300$ samples, higher-dimensional 20D Gaussian and higher-complexity sine wave datasets are much more challenging and are not solved using $N=3000$ samples with a signal-agnostic MLP architecture. \basealgo-sig and \metaalgo-sig detect $p<0.05$ statistically significant dependency on not only 1D and 20D Gaussian datasets where $x$ and $z$ have non-zero correlation, but also on the sine wave datasets where correlation between $x$ and $z$ is 0. This means that \basealgo-sig and \metaalgo-sig can be used for nonlinear dependency testing to complement linear correlation testing.

We study the effect of cross-validation meta-learning and task augmentation on 20D Gaussian with $\rho=0.3$ and $N=300$. Figure~\ref{fig:task_aug} plots performance of \metaalgo-vr over $N_M=3000$ meta iterations under combinations of task augmentations modes and number of adaptation iterations $N_O \in \{ 0,20\} $. 
Overall, task augmentation modes which involve axis flipping $m(\cdot )$ and permutation $P(\cdot)$ are the most successful. With $N_O=20$ steps of adaptation, task augmentation modes $P(\cdot)$, $m(P(\cdot))$ and $m(P(O(\cdot)))$ prevent overfitting and improves performance. The performance improvements of task augmentation is not simply from change in batch size, learning rate or number of optimization iterations, because meta-learning without task augmentation for both $N_O=0$ and $20$ could not outperform baseline. Meta-learning without task augmentation and with task augmentation but using only $O(\cdot)$ or $G(\cdot)$ result in overfitting. Task augmentation with $m(\cdot)$ or $m(P(O(G(\cdot))))$ prevent overfitting, but do not provide performance benefits, possibly because their complexity is insufficient or excessive for 20 adaptation steps. 
Further more, task augmentation with no adaptation ($N_O=0$) falls back to data augmentation, where samples from transformed distributions are directly used to learn $T_\theta(x,z)$. Data augmentation with $O(\cdot)$ outperforms no augmentation, but is unable to outperform baseline and suffer from overfitting. It shows that task augmentation provides improvements orthogonal to data augmentation.
\section{Application: fMRI Inter-subject correlation (ISC) analysis}
Humans use language to effectively transmit brain representations among conspecifics. For example, after witnessing an event in the world, a speaker may use verbal communication to evoke neural representations reflecting that event in a listener's brain \cite{Hasson_CogSci2012}. The efficacy of this transmission, in terms of listener comprehension, is predicted by speaker–listener neural synchrony and synchrony among listeners \cite{stephens2010speaker}. To date, most work has measured brain-to-brain synchrony by locating statistically significant inter-subject correlation (ISC); quantified as the Pearson product-moment correlation coefficient between response time series for corresponding voxels or regions of interest (ROIs) across individuals \cite{hasson2004intersubject,schippers2010mapping,silbert2014coupled}. Using \basealgo and \metaalgo for statistical dependency testing, we can extend ISC analysis to capture nonlinear and higher-order interactions in continuous fMRI responses. Specifically, given synchronized fMRI response frames in two brain regions $X$ and $Z$ across $K$ subjects ${X_{i}}, {Z_{i}},i=1,\ldots, K$ as random variables. We model the conditional mutual information $I(X_i;Z_j|i\neq j)$ as the MI form of pair-wise ISC analysis. By definition, $I(X_i;Z_j|i\neq j)$ first computes MI between activations $X_i$ and $Z_j$ from subjects $i$ and $j$ respectively, and then average across pairs of subjects $i \neq j$. It can be lower bounded using Eq.~\ref{eqn:tightBound} by learning a $T_\theta(x,z)$ shared across all subject pairs. 

\textbf{Dataset.} We study MI-based and correlation-based ISC on a fMRI story comprehension dataset~\cite{simony2016dynamic} with 40 participants listening to four spoken stories. Average story duration is 11 minutes. An fMRI frame with full brain coverage is captured at repetition time 1 TR $=$1.5 seconds with 2.5mm isotropic spatial resolution. We restricted our analysis to subsets of voxels defined using independent data from previous studies: functionally-defined masks of high ISC voxels (ISC; 3,800 voxels) and dorsal Default-Mode Network voxels (dDMN; 3,940 voxels) from \cite{simony2016dynamic} as well as 180 HCP-MMP1 multimodal cortex parcels from ~\cite{hcp_mmp1}. All masks were defined in MNI space. 

\textbf{Implementation.} We compare MI-based ISC using \basealgo and \metaalgo with correlation-based ISC using Pearson's correlation. \basealgo and \metaalgo setup follows Section~\secref{sec:expsetup}. The fMRI data were partitioned by subject into a train set of 20 subjects and a validation set of 20 different subjects. Residual 1D CNN is used instead of MLP as the encoder for studying temporal dependency. For Pearson's correlation, high-dimensional signals are reshaped to 1D for correlation analysis.

\textbf{Quantitative Results.}
We first study that for the fine grained HCM-MMP1 brain regions, which of them have $p<0.05$ statistically significant activities by MI and Pearson's correlation. Table~\ref{table:fmri_sig} shows the result. Overall, more regions have statistically significant correlation than dependency. This is expected because correlation requires less data to detect. But \metaalgo is able to find 6 brain regions that potentially have statistically significant dependency but lacks significant correlation. This shows that MI analysis can be used to complement correlation-based ISC analysis.

By considering temporal ISC over time, fMRI signals can be modeled with improved accuracy. In Table~\ref{table:fmri_classification} we apply \basealgo and \metaalgo with $L=10\text{TRs}$ (15s) sliding windows as random variables to study amount of information that can be extracted from ISC and dDMN masks. We use between-subject time-segment classification (BSC) for evaluation \cite{haxby2011common, guntupalli2016model}. Each fMRI scan is divided into $K$ non-overlapping $L=10\text{TRs}$ time segments. The BSC task is one versus rest retrieval: retrieve the corresponding time segment $z$ of an individual given a group of time segments ${x}$ excluding that individual, measured by top-1 accuracy. For retrieval score, $T_\theta(X,Z)$ is used for \basealgo and \metaalgo and $\rho(X,Z)$ is used for Pearson's correlation as a simple baseline. With CNN as encoder, \basealgo and \metaalgo model the signal better and achieve higher accuracy. Also. \metaalgo is able to extract 0.75 nats of MI from the ISC mask over $10\text{TRs}$ or 15s, which could potentially be improved by more samples and high frequency fMRI scans.

\begin{figure*}[t]
    \centering
    \caption{Top: Top contributing voxels in the learned $T_\theta(X,Z)$ by gradient magnitude $\mathbb{E}_X (\frac{\partial T}{\partial X_i})^2$. Auditory region is highlighted for ISC and GM masks (best in color). Bottom: Evaluation on the "Pie Man" dataset using the ISC mask showing our approach $T_\theta(X,Z)$ versus Pearson correlation over time in the one versus rest case averaged over 20 test subjects.}
    \includegraphics[width=\textwidth]{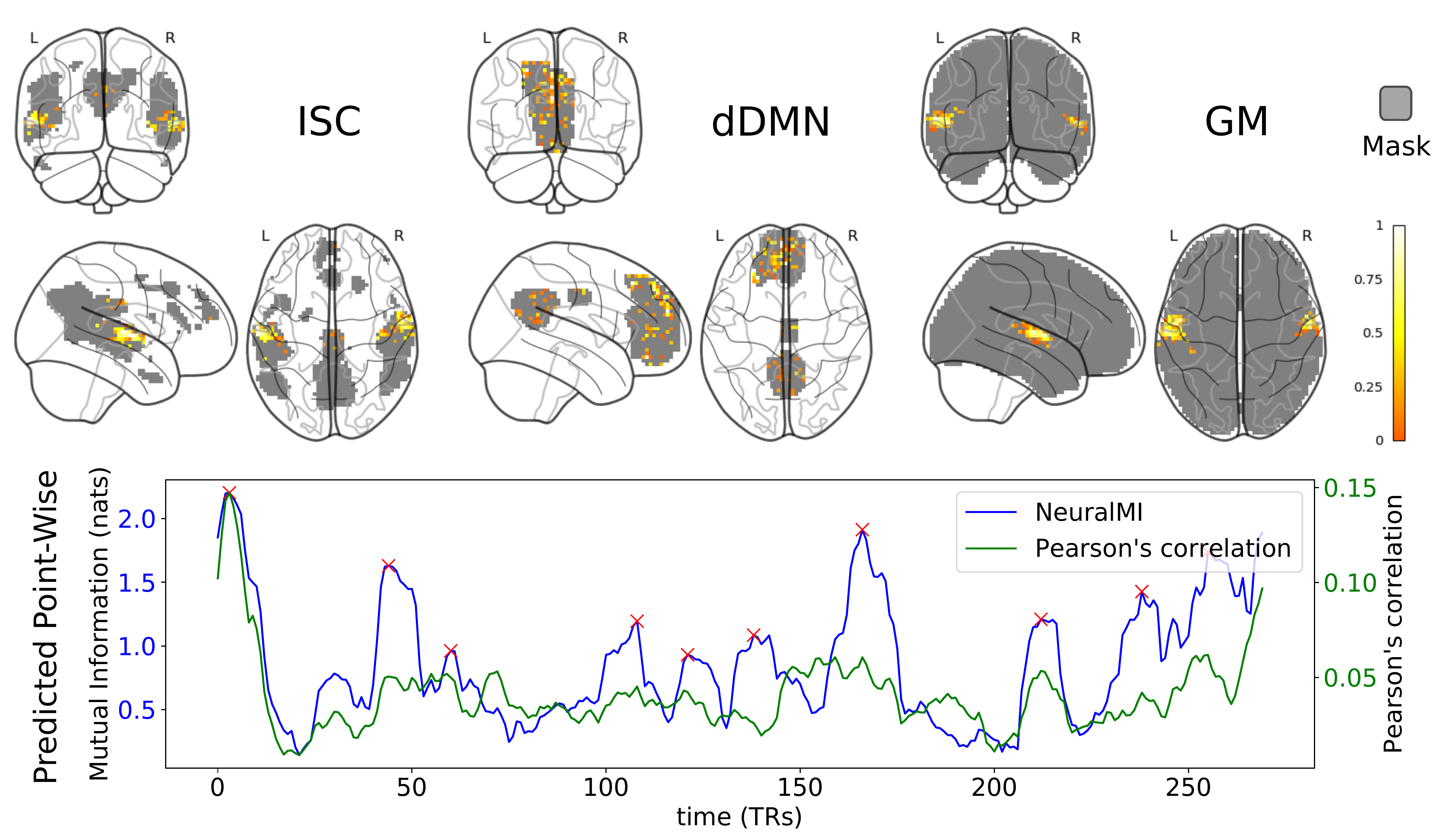}
    \label{fig:importance}
    \vspace{-20pt}
\end{figure*}

\textbf{Qualitative Results.} Fig.~\ref{fig:importance} (top) visualizes voxels that are important to $T_\theta(x,z)$ of the DEMINE model using their gradient magnitude variance for the ISC and dDMN masks, as well as an anatomically-defined Gray Matter (GM) mask. The DEMINE model focuses on auditory regions functionally important for perceiving the story stimulus.

Fig.~\ref{fig:importance} (bottom) plots the $T(x,z)$ and inter-subject Pearson correlations over time for "Pie Man" using the ISC mask and a sliding window size $L=10$, using the one vs rest scores averaged over all subjects. DEMINE yields more distinctive peaks.

We identify the peaks in \basealgo for "Pie Man" (with Pearson correlations) over time, then locate the story transcriptions in the $L=10 TRs$ (15 seconds) window corresponding to the peak:
\begin{itemize}
    \item 4: ``\ldots toiled for The Ram, uh, Fordham University’s student newspaper. And one day, I’m walking toward the campus center and out comes the elusive Dean McGowen, architect of a policy to replace traditionally \ldots"
    \item 45: "The Dean is covered with cream. So I give him a moment, then I say, `Dean McGowen, would you care to comment on this latest attack?' And he says, `Yes, I would care to comment. \ldots''
    \item 109: "\ldots which makes no sense. Fordham was a Catholic school and we all thought Latin was classy so, that’s what I used. And when I finished my story, I, I raced back to Dwyer and I showed it to him and he read it and he said \ldots"
    \item 122: "Few days later, I get a letter. I opened it up and it says, “Dear Jim, good story. Nice details. If you want to see me again in action, be on the steps of Duane Library \ldots"
    \item 139: "\ldots out comes student body president, Sheila Biel. And now, Sheila Biel was different from the rest of us flannel-shirt wearing, part-time-job working, Fordham students. Sheila was\ldots"
    \item 167: "Pie Man emerged from behind a late night library drop box, made his delivery, and fled away, crying, “Ego sum non una bestia.” And that’s what I reported in my story\ldots"
    \item 213: "\ldots that there was a question about whether she even knew if I existed. So I saw her there and made a mental note to do nothing about it, and then I went to the bar and ordered a drink, and I felt a, a tap on my shoulder. I turned around, and it was her.\ldots"
    \item 239: "And wasn’t I really Pie Man? Hadn’t I brought him into existence? Didn’t she only know about him because of me? But actually \ldots"
    \item 256: "I said, “Yes, Angela, I am Pie Man.' And she looked at me and she said, `Oh, good. I was hoping you’d say that \ldots"
\end{itemize}
We hypothesize that the scripts associated with the peaks may capture points when listeners pay more attention, resulting in the Signal-to-Noise Ratio (SNR) of fMRI scans being enhanced. 
\section{Conclusion}
%
%
%
We illustrated that a predictive view of the MI lower bounds coupled with meta-learning results in data-efficient variational MI estimators, \basealgo and \metaalgo, that are capable of performing statistical test of dependency. We also showed that our proposed task augmentation reduces overfitting and improves generalization in meta-learning. 
We successfully applied MI estimation to real world, data scarce, fMRI datasets. Our results suggest a greater avenue of using neural networks and meta-learning to improve MI analysis and applying neural network-based information theory tools to enhance the analysis of information processing in the brain.
Model-agnostic, high-confidence, MI lower bound estimation approaches -- including \textit{MINE}, \basealgo and \metaalgo -- are limited to estimating small MI lower bounds up to $O(\log N )$ as pointed out in~\cite{McAllester2018}, where $N$ is the number of samples. In real fMRI datasets, however, strong dependency is rare and existing MI estimation tools are limited more by their ability to accurately characterize the dependency. Nevertheless, when quantitatively measuring strong dependency, cross-entropy~\cite{McAllester2018} or model-based quantities, alternatives to MI, such as correlation or CCA, may be measured with high confidence.

\section*{Acknowledgments}
This work is funded by DARPA FA8750-18-C-0213. The views, opinions, and/or conclusions contained in this paper are those of the author and should not be interpreted as representing the official views or policies, either expressed or implied of the DARPA or the DoD.

\clearpage
\bibliography{neuralmi}

\begin{thebibliography}{10}\itemsep=-1pt

\bibitem{Agakov2004}
D.~B.~F. Agakov.
\newblock The {IM} algorithm: a variational approach to information
  maximization.
\newblock {\em Advances in Neural Information Processing Systems}, 16:201,
  2004.

\bibitem{Ahmad1976}
I.~Ahmad and P.-E. Lin.
\newblock A nonparametric estimation of the entropy for absolutely continuous
  distributions (corresp.).
\newblock {\em IEEE Transactions on Information Theory}, 22(3):372--375, 1976.

\bibitem{Avants_2008}
B.~B. Avants, C.~L. Epstein, M.~Grossman, and J.~C. Gee.
\newblock Symmetric diffeomorphic image registration with cross-correlation:
  evaluating automated labeling of elderly and neurodegenerative brain.
\newblock {\em Medical Image Analysis}, 12(1):26--41, 2008.

\bibitem{behzadi2007component}
Y.~Behzadi, K.~Restom, J.~Liau, and T.~T. Liu.
\newblock {A component based noise correction method (CompCor) for BOLD and
  perfusion based fMRI}.
\newblock {\em NeuroImage}, 37(1):90--101, 2007.

\bibitem{MINE}
M.~I. Belghazi, A.~Baratin, S.~Rajeshwar, S.~Ozair, Y.~Bengio, D.~Hjelm, and
  A.~Courville.
\newblock Mutual information neural estimation.
\newblock In {\em International Conference on Machine Learning}, pages
  530--539, 2018.

\bibitem{hyperopt}
J.~Bergstra, D.~Yamins, and D.~D. Cox.
\newblock Making a science of model search: Hyperparameter optimization in
  hundreds of dimensions for vision architectures.
\newblock 2013.

\bibitem{Cox_1996}
R.~W. Cox.
\newblock {AFNI}: software for analysis and visualization of functional
  magnetic resonance neuroimages.
\newblock {\em Computers and Biomedical research}, 29(3):162--173, 1996.

\bibitem{Black}
C.~Daniel.
\newblock I knew you were black.
\newblock \url{https://themoth.org/stories/i-knew-you-were-black}, 2018.
\newblock Accessed: 2018-10-12.

\bibitem{Esteban_2018}
O.~Esteban, C.~Markiewicz, R.~W. Blair, C.~Moodie, A.~I. Isik,
  A.~Erramuzpe~Aliaga, J.~Kent, M.~Goncalves, E.~DuPre, M.~Snyder, H.~Oya,
  S.~Ghosh, J.~Wright, J.~Durnez, R.~Poldrack, and K.~J. Gorgolewski.
\newblock {FMRIPrep}: a robust preprocessing pipeline for functional {MRI}.
\newblock {\em bioRxiv}, 2018.

\bibitem{MAML}
C.~Finn, P.~Abbeel, and S.~Levine.
\newblock Model-agnostic meta-learning for fast adaptation of deep networks.
\newblock In {\em Proceedings of the 34th International Conference on Machine
  Learning, {ICML} 2017, Sydney, NSW, Australia, 6-11 August 2017}, pages
  1126--1135, 2017.

\bibitem{Finn2018}
C.~Finn, K.~Xu, and S.~Levine.
\newblock Probabilistic model-agnostic meta-learning.
\newblock In {\em Advances in Neural Information Processing Systems}, pages
  9537--9548, 2018.

\bibitem{Finn2017_imitation}
C.~Finn, T.~Yu, T.~Zhang, P.~Abbeel, and S.~Levine.
\newblock One-shot visual imitation learning via meta-learning.
\newblock In {\em Conference on Robot Learning}, pages 357--368, 2017.

\bibitem{Fonov_2009}
V.~S. Fonov, A.~C. Evans, R.~C. McKinstry, C.~Almli, and D.~Collins.
\newblock Unbiased nonlinear average age-appropriate brain templates from birth
  to adulthood.
\newblock {\em NeuroImage}, (47):S102, 2009.

\bibitem{Forgot}
N.~Gaiman.
\newblock The man who forgot ray bradbury.
\newblock
  \url{https://soundcloud.com/neilgaiman/the-man-who-forgot-ray-bradbury},
  2018.
\newblock Accessed: 2018-10-12.

\bibitem{Gao2017}
W.~Gao, S.~Kannan, S.~Oh, and P.~Viswanath.
\newblock Estimating mutual information for discrete-continuous mixtures.
\newblock In {\em Advances in Neural Information Processing Systems}, pages
  5986--5997, 2017.

\bibitem{Gao2018}
W.~Gao, S.~Oh, and P.~Viswanath.
\newblock Demystifying fixed $ k $-nearest neighbor information estimators.
\newblock {\em IEEE Transactions on Information Theory}, 64(8):5629--5661,
  2018.

\bibitem{hcp_mmp1}
M.~F. Glasser, T.~S. Coalson, E.~C. Robinson, C.~D. Hacker, J.~Harwell,
  E.~Yacoub, K.~Ugurbil, J.~Andersson, C.~F. Beckmann, M.~Jenkinson, et~al.
\newblock A multi-modal parcellation of human cerebral cortex.
\newblock {\em Nature}, 536(7615):171, 2016.

\bibitem{xavier}
X.~Glorot and Y.~Bengio.
\newblock Understanding the difficulty of training deep feedforward neural
  networks.
\newblock In {\em In Proceedings of the International Conference on Artificial
  Intelligence and Statistics (AISTATS’10). Society for Artificial
  Intelligence and Statistics}, 2010.

\bibitem{Gorgolewski_2011}
K.~Gorgolewski, C.~Burns, C.~Madison, D.~Clark, Y.~Halchenko, M.~Waskom, and
  S.~Ghosh.
\newblock Nipype: a flexible, lightweight and extensible neuroimaging data
  processing framework in python.
\newblock {\em Frontiers in Neuroinformatics}, 5:13, 2011.

\bibitem{gorgolewski2016brain}
K.~J. Gorgolewski, T.~Auer, V.~D. Calhoun, R.~C. Craddock, S.~Das, E.~P. Duff,
  G.~Flandin, S.~S. Ghosh, T.~Glatard, Y.~O. Halchenko, et~al.
\newblock The brain imaging data structure, a format for organizing and
  describing outputs of neuroimaging experiments.
\newblock {\em Scientific Data}, 3:160044, 2016.

\bibitem{greve2009accurate}
D.~N. Greve and B.~Fischl.
\newblock Accurate and robust brain image alignment using boundary-based
  registration.
\newblock {\em NeuroImage}, 48(1):63--72, 2009.

\bibitem{guntupalli2016model}
J.~S. Guntupalli, M.~Hanke, Y.~O. Halchenko, A.~C. Connolly, P.~J. Ramadge, and
  J.~V. Haxby.
\newblock A model of representational spaces in human cortex.
\newblock {\em Cerebral Cortex}, 26(6):2919--2934, 2016.

\bibitem{Hasson_CogSci2012}
U.~Hasson, A.~A. Ghazanfar, B.~Galantucci, S.~Garrod, and C.~Keysers.
\newblock Brain-to-brain coupling: a mechanism for creating and sharing a
  social world.
\newblock {\em Trends in cognitive sciences}, 16(2):114--121, 2012.

\bibitem{hasson2004intersubject}
U.~Hasson, Y.~Nir, I.~Levy, G.~Fuhrmann, and R.~Malach.
\newblock Intersubject synchronization of cortical activity during natural
  vision.
\newblock {\em Science}, 303(5664):1634--1640, 2004.

\bibitem{haxby2011common}
J.~V. Haxby, J.~S. Guntupalli, A.~C. Connolly, Y.~O. Halchenko, B.~R. Conroy,
  M.~I. Gobbini, M.~Hanke, and P.~J. Ramadge.
\newblock A common, high-dimensional model of the representational space in
  human ventral temporal cortex.
\newblock {\em Neuron}, 72(2):404--416, 2011.

\bibitem{jenkinson2002improved}
M.~Jenkinson, P.~Bannister, M.~Brady, and S.~Smith.
\newblock Improved optimization for the robust and accurate linear registration
  and motion correction of brain images.
\newblock {\em NeuroImage}, 17(2):825--841, 2002.

\bibitem{Kim2018}
T.~Kim, J.~Yoon, O.~Dia, S.~Kim, Y.~Bengio, and S.~Ahn.
\newblock Bayesian model-agnostic meta-learning.
\newblock {\em arXiv preprint arXiv:1806.03836}, 2018.

\bibitem{adam}
D.~P. Kingma and J.~Ba.
\newblock Adam: A method for stochastic optimization.
\newblock {\em arXiv preprint arXiv:1412.6980}, 2014.

\bibitem{Kraskov2004}
A.~Kraskov, H.~Stogbauer, and P.~Grassberger.
\newblock Estimating mutual information.
\newblock {\em Physical review E}, 2004.

\bibitem{Maclaurin2015}
D.~Maclaurin, D.~Duvenaud, and R.~Adams.
\newblock Gradient-based hyperparameter optimization through reversible
  learning.
\newblock In {\em International Conference on Machine Learning}, pages
  2113--2122, 2015.

\bibitem{McAllester2018}
D.~McAllester and K.~Statos.
\newblock Formal limitations on the measurement of mutual information.
\newblock {\em arXiv preprint arXiv:1811.04251}, 2018.

\bibitem{PieMan}
J.~O'Grady.
\newblock Pie {Man}.
\newblock \url{https://themoth.org/stories/pie-man}, 2018.
\newblock Accessed: 2018-10-12.

\bibitem{Bronx}
J.~O'Grady.
\newblock Running from the {Bronx}.
\newblock
  \url{https://soundcloud.com/the-story-collider/jim-ogrady-running-from-the},
  2018.
\newblock Accessed: 2018-10-12.

\bibitem{Pham2018}
H.~Pham, M.~Guan, B.~Zoph, Q.~Le, and J.~Dean.
\newblock Efficient neural architecture search via parameter sharing.
\newblock In {\em International Conference on Machine Learning}, pages
  4092--4101, 2018.

\bibitem{Poole_2018}
B.~Poole, S.~Ozair, A.~van~den Oord, A.~A. Alemi, and G.~Tucker.
\newblock On variational lower bounds of mutual information.
\newblock In {\em Bayesian Deep Learning Workshop, NeurIPSW}, 2018.

\bibitem{power2014methods}
J.~D. Power, A.~Mitra, T.~O. Laumann, A.~Z. Snyder, B.~L. Schlaggar, and S.~E.
  Petersen.
\newblock Methods to detect, characterize, and remove motion artifact in
  resting state {fMRI}.
\newblock {\em NeuroImage}, 84:320--341, 2014.

\bibitem{es}
T.~Salimans, J.~Ho, X.~Chen, S.~Sidor, and I.~Sutskever.
\newblock Evolution strategies as a scalable alternative to reinforcement
  learning.
\newblock {\em arXiv preprint arXiv:1703.03864}, 2017.

\bibitem{schippers2010mapping}
M.~B. Schippers, A.~Roebroeck, R.~Renken, L.~Nanetti, and C.~Keysers.
\newblock Mapping the information flow from one brain to another during
  gestural communication.
\newblock {\em Proceedings of the National Academy of Sciences}, page
  201001791, 2010.

\bibitem{pepg}
F.~Sehnke, C.~Osendorfer, T.~R{\"u}ckstie{\ss}, A.~Graves, J.~Peters, and
  J.~Schmidhuber.
\newblock Parameter-exploring policy gradients.
\newblock {\em Neural Networks}, 23(4):551--559, 2010.

\bibitem{silbert2014coupled}
L.~J. Silbert, C.~J. Honey, E.~Simony, D.~Poeppel, and U.~Hasson.
\newblock Coupled neural systems underlie the production and comprehension of
  naturalistic narrative speech.
\newblock {\em Proceedings of the National Academy of Sciences},
  111(43):E4687--E4696, 2014.

\bibitem{simony2016dynamic}
E.~Simony, C.~J. Honey, J.~Chen, O.~Lositsky, Y.~Yeshurun, A.~Wiesel, and
  U.~Hasson.
\newblock Dynamic reconfiguration of the default mode network during narrative
  comprehension.
\newblock {\em Nature Communications}, 7:12141, 2016.

\bibitem{Snell2017}
J.~Snell, K.~Swersky, and R.~Zemel.
\newblock Prototypical networks for few-shot learning.
\newblock In {\em Advances in Neural Information Processing Systems}, pages
  4077--4087, 2017.

\bibitem{stephens2010speaker}
G.~J. Stephens, L.~J. Silbert, and U.~Hasson.
\newblock Speaker--listener neural coupling underlies successful communication.
\newblock {\em Proceedings of the National Academy of Sciences},
  107(32):14425--14430, 2010.

\bibitem{treiber2016characterization}
J.~M. Treiber, N.~S. White, T.~C. Steed, H.~Bartsch, D.~Holland, N.~Farid,
  C.~R. McDonald, B.~S. Carter, A.~M. Dale, and C.~C. Chen.
\newblock Characterization and correction of geometric distortions in 814
  diffusion weighted images.
\newblock {\em PLOS ONE}, 11(3):e0152472, 2016.

\bibitem{Tustison_2010}
N.~J. Tustison, B.~B. Avants, P.~A. Cook, Y.~Zheng, A.~Egan, P.~A. Yushkevich,
  and J.~C. Gee.
\newblock N4itk: improved n3 bias correction.
\newblock {\em IEEE Transactions on Medical Imaging}, 29(6):1310--1320, June
  2010.

\bibitem{Vinyals2016}
O.~Vinyals, C.~Blundell, T.~Lillicrap, D.~Wierstra, et~al.
\newblock Matching networks for one shot learning.
\newblock In {\em Advances in neural information processing systems}, pages
  3630--3638, 2016.

\bibitem{wang2017evaluation}
S.~Wang, D.~J. Peterson, J.~C. Gatenby, W.~Li, T.~J. Grabowski, and T.~M.
  Madhyastha.
\newblock Evaluation of field map and nonlinear registration methods for
  correction of susceptibility artifacts in diffusion mri.
\newblock {\em Frontiers in Neuroinformatics}, 11:17, 2017.

\bibitem{Zhang_2016}
C.~Zhang, S.~Bengio, M.~Hardt, B.~Recht, and O.~Vinyals.
\newblock Understanding deep learning requires rethinking generalization.
\newblock {\em arXiv preprint arXiv:1611.03530}, 2016.

\bibitem{Zhang_2001}
Y.~Zhang, M.~Brady, and S.~Smith.
\newblock Segmentation of brain {MR} images through a hidden markov random
  field model and the expectation-maximization algorithm.
\newblock {\em IEEE Transactions on Medical Imaging}, 20(1):45--57, 2001.

\end{thebibliography}


\begin{thebibliography}{10}\itemsep=-1pt

\bibitem{Avants_2008}
B.~B. Avants, C.~L. Epstein, M.~Grossman, and J.~C. Gee.
\newblock Symmetric diffeomorphic image registration with cross-correlation:
  evaluating automated labeling of elderly and neurodegenerative brain.
\newblock {\em Medical Image Analysis}, 12(1):26--41, 2008.

\bibitem{behzadi2007component}
Y.~Behzadi, K.~Restom, J.~Liau, and T.~T. Liu.
\newblock {A component based noise correction method (CompCor) for BOLD and
  perfusion based fMRI}.
\newblock {\em NeuroImage}, 37(1):90--101, 2007.

\bibitem{Cox_1996}
R.~W. Cox.
\newblock {AFNI}: software for analysis and visualization of functional
  magnetic resonance neuroimages.
\newblock {\em Computers and Biomedical research}, 29(3):162--173, 1996.

\bibitem{Black}
C.~Daniel.
\newblock I knew you were black.
\newblock \url{https://themoth.org/stories/i-knew-you-were-black}, 2018.
\newblock Accessed: 2018-10-12.

\bibitem{Esteban_2018}
O.~Esteban, C.~Markiewicz, R.~W. Blair, C.~Moodie, A.~I. Isik,
  A.~Erramuzpe~Aliaga, J.~Kent, M.~Goncalves, E.~DuPre, M.~Snyder, H.~Oya,
  S.~Ghosh, J.~Wright, J.~Durnez, R.~Poldrack, and K.~J. Gorgolewski.
\newblock {FMRIPrep}: a robust preprocessing pipeline for functional {MRI}.
\newblock {\em bioRxiv}, 2018.

\bibitem{Fonov_2009}
V.~S. Fonov, A.~C. Evans, R.~C. McKinstry, C.~Almli, and D.~Collins.
\newblock Unbiased nonlinear average age-appropriate brain templates from birth
  to adulthood.
\newblock {\em NeuroImage}, (47):S102, 2009.

\bibitem{Forgot}
N.~Gaiman.
\newblock The man who forgot ray bradbury.
\newblock
  \url{https://soundcloud.com/neilgaiman/the-man-who-forgot-ray-bradbury},
  2018.
\newblock Accessed: 2018-10-12.

\bibitem{xavier}
X.~Glorot and Y.~Bengio.
\newblock Understanding the difficulty of training deep feedforward neural
  networks.
\newblock In {\em In Proceedings of the International Conference on Artificial
  Intelligence and Statistics (AISTATS’10). Society for Artificial
  Intelligence and Statistics}, 2010.

\bibitem{Gorgolewski_2011}
K.~Gorgolewski, C.~Burns, C.~Madison, D.~Clark, Y.~Halchenko, M.~Waskom, and
  S.~Ghosh.
\newblock Nipype: a flexible, lightweight and extensible neuroimaging data
  processing framework in python.
\newblock {\em Frontiers in Neuroinformatics}, 5:13, 2011.

\bibitem{gorgolewski2016brain}
K.~J. Gorgolewski, T.~Auer, V.~D. Calhoun, R.~C. Craddock, S.~Das, E.~P. Duff,
  G.~Flandin, S.~S. Ghosh, T.~Glatard, Y.~O. Halchenko, et~al.
\newblock The brain imaging data structure, a format for organizing and
  describing outputs of neuroimaging experiments.
\newblock {\em Scientific Data}, 3:160044, 2016.

\bibitem{greve2009accurate}
D.~N. Greve and B.~Fischl.
\newblock Accurate and robust brain image alignment using boundary-based
  registration.
\newblock {\em NeuroImage}, 48(1):63--72, 2009.

\bibitem{jenkinson2002improved}
M.~Jenkinson, P.~Bannister, M.~Brady, and S.~Smith.
\newblock Improved optimization for the robust and accurate linear registration
  and motion correction of brain images.
\newblock {\em NeuroImage}, 17(2):825--841, 2002.

\bibitem{adam}
D.~P. Kingma and J.~Ba.
\newblock Adam: A method for stochastic optimization.
\newblock {\em arXiv preprint arXiv:1412.6980}, 2014.

\bibitem{PieMan}
J.~O'Grady.
\newblock Pie {Man}.
\newblock \url{https://themoth.org/stories/pie-man}, 2018.
\newblock Accessed: 2018-10-12.

\bibitem{Bronx}
J.~O'Grady.
\newblock Running from the {Bronx}.
\newblock
  \url{https://soundcloud.com/the-story-collider/jim-ogrady-running-from-the},
  2018.
\newblock Accessed: 2018-10-12.

\bibitem{power2014methods}
J.~D. Power, A.~Mitra, T.~O. Laumann, A.~Z. Snyder, B.~L. Schlaggar, and S.~E.
  Petersen.
\newblock Methods to detect, characterize, and remove motion artifact in
  resting state {fMRI}.
\newblock {\em NeuroImage}, 84:320--341, 2014.

\bibitem{treiber2016characterization}
J.~M. Treiber, N.~S. White, T.~C. Steed, H.~Bartsch, D.~Holland, N.~Farid,
  C.~R. McDonald, B.~S. Carter, A.~M. Dale, and C.~C. Chen.
\newblock Characterization and correction of geometric distortions in 814
  diffusion weighted images.
\newblock {\em PLOS ONE}, 11(3):e0152472, 2016.

\bibitem{Tustison_2010}
N.~J. Tustison, B.~B. Avants, P.~A. Cook, Y.~Zheng, A.~Egan, P.~A. Yushkevich,
  and J.~C. Gee.
\newblock N4itk: improved n3 bias correction.
\newblock {\em IEEE Transactions on Medical Imaging}, 29(6):1310--1320, June
  2010.

\bibitem{wang2017evaluation}
S.~Wang, D.~J. Peterson, J.~C. Gatenby, W.~Li, T.~J. Grabowski, and T.~M.
  Madhyastha.
\newblock Evaluation of field map and nonlinear registration methods for
  correction of susceptibility artifacts in diffusion mri.
\newblock {\em Frontiers in Neuroinformatics}, 11:17, 2017.

\bibitem{Zhang_2001}
Y.~Zhang, M.~Brady, and S.~Smith.
\newblock Segmentation of brain {MR} images through a hidden markov random
  field model and the expectation-maximization algorithm.
\newblock {\em IEEE Transactions on Medical Imaging}, 20(1):45--57, 2001.

\end{thebibliography}
\bibliographystyle{ieee}
\clearpage
\appendix
\section{Additional Details about the fMRI Dataset}
The dataset we use~\cite{simony2016dynamic}, contains 40 participants (mean age = 23.3 years, SD = 8.9, range: 18–53; 27 female) recruited to listen to four spoken stories\footnote{Two of the stories were told by a professional storyteller undergoing an fMRI scan; however, fMRI data for the speaker were not analyzed for the present work due to the head motion induced by speech production.}\footnote{The study was conducted in compliance with the Institutional Review Board of the University}. The stories were renditions of “Pie Man” and “Running from the Bronx” by Jim O’Grady \cite{PieMan, Bronx}, “The Man Who Forgot Ray Bradbury” by Neil Gaiman \cite{Forgot}, and “I Knew You Were Black” by Carol Daniel \cite{Black}; story durations were ~7, 9, 14, and 13 minutes, respectively. After scanning, participants completed a questionnaire comprising 25–30 questions per story intended to measure narrative comprehension. The questionnaires included multiple choice, True/False, and fill-in-the-blank questions, as well as four additional subjective ratings per story. Functional and structural images were acquired using a 3T Siemens Prisma with a 64-channel head coil (see Section~\secref{sec:dataset_collection} for additional details). Briefly, functional images were acquired in an interleaved fashion using gradient-echo echo-planar imaging with a multiband acceleration factor of 3 (TR/TE = 1500/31 ms, resolution = 2.5 mm isotropic voxels, full brain coverage).

All fMRI data were formatted according to the Brain Imaging Data Structure (BIDS) standard \cite{gorgolewski2016brain} and preprocessed using fMRIPrep \cite{Esteban_2018} (see Section~\secref{sec:dataset_processing} for additional details). Functional data were corrected for slice timing, head motion, and susceptibility distortion, and normalized to MNI space using nonlinear registration. Nuisance variables comprising head motion parameters, framewise displacement, linear and quadratic trends, sine/cosine bases for high-pass filtering (0.007 Hz), and six principal component time series from cerebrospinal fluid (CSF) and white matter were regressed out of the signal using AFNI \cite{Cox_1996}.

The fMRI data comprise \dimension{\FMRI}{\FMRIdimFlat} for each subject, where \xV represents the flattened and masked voxel space and \T represents the number of samples (TRs) during auditory stimulus presentation.

\subsection{Additional Details on Dataset Collection}
\label{sec:dataset_collection}
Functional and structural images were acquired using a 3T Siemens Magnetom Prisma with a 64-channel head coil. Functional, blood-oxygenation-level-dependent (BOLD) images were acquired in an interleaved fashion using gradient-echo echo-planar imaging with pre-scan normalization, fat suppression, a multiband acceleration factor of 3, and no in-plane acceleration: TR/TE = 1500/31 ms, flip angle = 67$^{\circ}$, bandwidth = 2480 Hz/Px, resolution = 2.5 mm3 isotropic voxels, matrix size = 96 x 96, FoV = 240 x 240 mm, 48 axial slices with roughly full brain coverage and no gap, anterior–posterior phase encoding. At the beginning of each scanning session, a T1-weighted structural scan was acquired using a high-resolution single-shot MPRAGE sequence with an in-plane acceleration factor of 2 using GRAPPA: TR/TE/TI = 2530/3.3/1100 ms, flip angle = 7$^{\circ}$, resolution = 1.0 x 1.0 x 1.0 mm voxels, matrix size = 256 x 256, FoV = 256 x 256 x 176 mm, 176 sagittal slices, ascending acquisition, anterior–posterior phase encoding, no fat suppression, 5 min 53 s total acquisition time. At the end of each scanning session a T2-weighted structural scan was acquired using the same acquisition parameters and geometry as the T1-weighted structural image: TR/TE = 3200/428 ms, 4 min 40 s total acquisition time. A field map was acquired at the beginning of each scanning session, but was not used in subsequent analyses.

\subsection{Additional Details on Dataset Preprocessing}\label{sec:dataset_processing}
Preprocessing was performed using fMRIPrep \cite{Esteban_2018}, a Nipype \cite{Gorgolewski_2011} based tool. T1-weighted images were corrected for intensity non-uniformity using N4 bias field correction \cite{Tustison_2010} and skull-stripped using ANTs \cite{Avants_2008}. Nonlinear spatial normalization to the ICBM 152 Nonlinear Asymmetrical template version 2009c \cite{Fonov_2009} was performed using ANTs. Brain tissue segmentation cerebrospinal fluid (CSF), white matter, and gray matter was  was performed using FSL's FAST \cite{Zhang_2001}. Functional images were slice timing corrected using AFNI's 3dTshift \cite{Cox_1996} and corrected for head motion using FSL's MCFLIRT \cite{jenkinson2002improved}. "Fieldmap-less" distortion correction was performed by  co-registering each subject's functional image to that subject's intensity-inverted T1-weighted image \cite{wang2017evaluation} constrained with an average field map template \cite{treiber2016characterization}. This was followed by co-registration to the corresponding T1-weighted image using FreeSurfer's boundary-based registration \cite{greve2009accurate} with 9 degrees of freedom. Motion correcting transformations, field distortion correcting warp, BOLD-to-T1 transformation and T1-to-template (MNI) warp were concatenated and applied in a single step with Lanczos interpolation using ANTs. Physiological noise regressors were extracted applying aCompCor \cite{behzadi2007component}. Six principal component time series were calculated within the intersection of the subcortical mask and the union of CSF and WM masks calculated in T1w space, after their projection to the native space of each functional run. Framewise displacement \cite{power2014methods} was calculated for each functional run. Functional images were downsampled to 3 mm resolution. Nuisance variables comprising six head motion parameters (and their derivatives), framewise displacement, linear and quadratic trends, sine/cosine bases for high-pass filtering (0.007 Hz cutoff), and six principal component time series from an anatomically-defined mask of cerebrospinal fluid (CSF) and white matter were regressed out of the signal using AFNI’s 3dTproject \cite{Cox_1996}. Functional response time series were z-scored for each voxel.

\end{document}